\documentclass[11pt]{article}
\usepackage{fullpage}
\usepackage{supertabular}

\usepackage{amssymb,url,graphicx,dsfont} 
\usepackage{float}
\floatstyle{ruled}
\newfloat{algo}{thp}{lop}
\floatname{algo}{Algorithm}

\newtheorem{remark}{Remark}

\newenvironment{keywords}{{\bf Keywords:}}

\def\myname{}
\def\email{\\ E-mail:}
\def\myaddr{}

\begin{document}

\def\thetaspace{\mathds{N}}
\def\etaspace{\mathds{M}}
\def\lambdaspace{\mathds{L}}

\def\IS{\mathrm{IS}}
\def\var{\mathrm{var}}
\def\C{\mathcal{C}}
\def\P{\mathcal{P}}
\def\inner#1#2{\langle #1,#2\rangle}
\def\Inner#1#2{\left\langle #1,#2\right\rangle}
\def\dx{\mathrm{d}x}
\def\argmax{\mathrm{argmax}}
\def\argmin{\mathrm{argmin}}
\def\opt{{\mathrm{opt}}}
\def\V{\mathcal{V}} 
\def\X{\mathcal{X}} 
\def\doi#1{\url{doi:#1}}
\def\KL{\mathrm{KL}}
\def\tr{\mathrm{tr}}
\def\Y{\mathcal{Y}}
\def\kmeans{\mathrm{kmeans}}
\def\MC{\mathrm{MC}}

\title{$k$-MLE: A fast algorithm for learning statistical mixture models\thanks{Research performed during the January-June 2011 period. 
A preliminary shorter version appeared in  IEEE International Conference on Acoustics,
Speech, and Signal Processing (ICASSP) 2012.}
}

\author{\myname Frank Nielsen \\
       \myaddr Sony Computer Science Laboratories, Inc\\
       3-14-13 Higashi Gotanda\\
       141-0022 Shinagawa-Ku, Tokyo, Japan\\
       \email Frank.Nielsen@acm.org\\ 
       }

\date{June 2011 (revised March 2012)}

\maketitle

\begin{abstract}
We describe $k$-MLE, a fast and efficient local search algorithm for learning finite statistical mixtures of exponential families such as Gaussian mixture models. Mixture models are traditionally learned using the expectation-maximization (EM) soft clustering technique that monotonically increases the incomplete (expected complete) likelihood.
Given prescribed mixture weights, the hard clustering $k$-MLE algorithm iteratively assigns data to the most likely weighted component and update the component models using Maximum Likelihood Estimators (MLEs). 
Using the duality between exponential families and Bregman divergences, we prove that the local convergence of the complete likelihood of $k$-MLE follows directly from the convergence of a dual additively weighted Bregman hard clustering. 
The inner loop of $k$-MLE can be implemented using any $k$-means heuristic like the celebrated Lloyd's batched or Hartigan's greedy swap updates.
We then show how to update the mixture weights by minimizing a cross-entropy criterion that implies to update weights by taking the relative proportion of cluster points, and reiterate the mixture parameter update and mixture weight update processes until convergence.
Hard EM is interpreted as a special case of $k$-MLE when both the component update and the weight update are performed successively in the inner loop. 
To initialize $k$-MLE, we propose $k$-MLE++, a careful initialization of $k$-MLE guaranteeing probabilistically a global bound on the best possible  complete likelihood. 
\end{abstract}

\begin{keywords}
exponential families, mixtures, Bregman divergences, expectation-maximization (EM), $k$-means loss function, Lloyd's $k$-means, Hartigan and Wong's $k$-means, hard EM, sparse EM.
\end{keywords}

\sloppy

\section{Introduction}

\subsection{Statistical mixture models}

A statistical mixture model~\cite{FiniteMixtureModels:2000} $M\sim m$ with $k\in\mathbb{N}$ weighted components has underlying probability distribution:

\begin{equation}
m(x|w,\theta) = \sum_{i=1}^k  w_i p(x|\theta_i),
\end{equation}
with $w=(w_1, ..., w_k)$ and $\theta=(\theta_1, ..., \theta_k)$ denoting the mixture parameters:
The $w_i$'s are positive weights summing up to one, 
and the $\theta_i$'s denote the individual component parameters.
(Appendix~\ref{sec:notations} summarizes the notations used throughout the paper.)

Mixture models of $d$-dimensional Gaussians\footnote{Also called MultiVariate Normals (MVNs) in software packages.} are the most often used statistical mixtures~\cite{FiniteMixtureModels:2000}.
In that case, each component distribution $N(\mu_i,\Sigma_i)$ is parameterized by a mean vector $\mu_i\in\mathbb{R}^d$ and a covariance matrix $\Sigma_i\succ 0$ that is symmetric and positive definite.
That is, $\theta_i=(\mu_i,\Sigma_i)$. 
The Gaussian distribution has the following probability density defined on the support $\mathbb{X}=\mathbb{R}^d$:

\begin{equation}\label{eq:mvn}
p(x;\mu_i,\Sigma_i)=\frac{1}{(2\pi)^{\frac{d}{2}}\sqrt{|\Sigma_i|}}e^{-\frac{1}{2} M_{\Sigma_i^{-1}}(x-\mu_i,x-\mu_i)},
\end{equation}
where $M_Q$ denotes the squared Mahalanobis distance~\cite{bvd-2010} 
\begin{equation}
M_{Q}(x,y)=(x-y)^T Q (x-y),
\end{equation}
defined for a symmetric positive definite matrix $Q\succ 0$ ($Q_i=\Sigma_i^{-1}$, the precision matrix).

To draw a random variate from a Gaussian mixture model (GMM) with $k$ components, we first draw a multinomial variate $z\in\{1, ...,k\}$, and then sample a Gaussian variate from $N(\mu_z,\Sigma_z)$. A multivariate normal variate $x$ is drawn from the chosen component $N(\mu,\Sigma)$ as follows:
First, we consider the Cholesky decomposition of the covariance matrix: $\Sigma=CC^T$, and take a $d$-dimensional vector with coordinates being random standard normal variates: $y=[y_1\ ...\ y_d]^T$ with $y_i=\sqrt{-2\log u_1}\cos(2\pi u_2)$ (for $u_1$ and $u_2$ uniform random variates in $[0,1)$).
Finally, we assemble the Gaussian variate $x$ as $x=\mu+Cy$.
This drawing process emphasizes that sampling a statistical mixture is a {\it doubly stochastic process} by essence: First, we sample a multinomial law for choosing the component, and then we sample the variate from the selected component.

\begin{figure}
\centering
(a)\includegraphics[bb=0 0 512 512,width=0.25\textwidth]{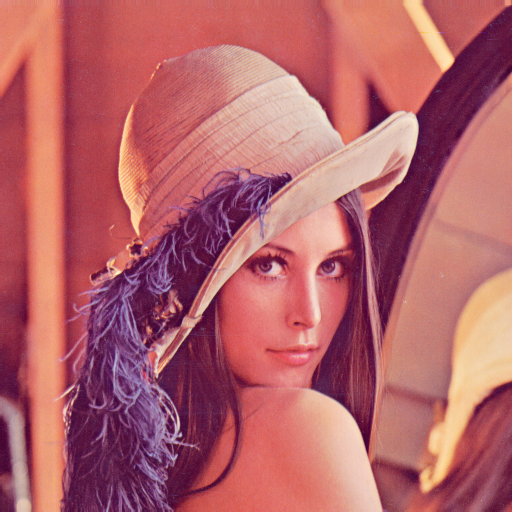}
(b)\includegraphics[bb=0 0 800 600,width=0.25\textwidth, height=0.25\textwidth]{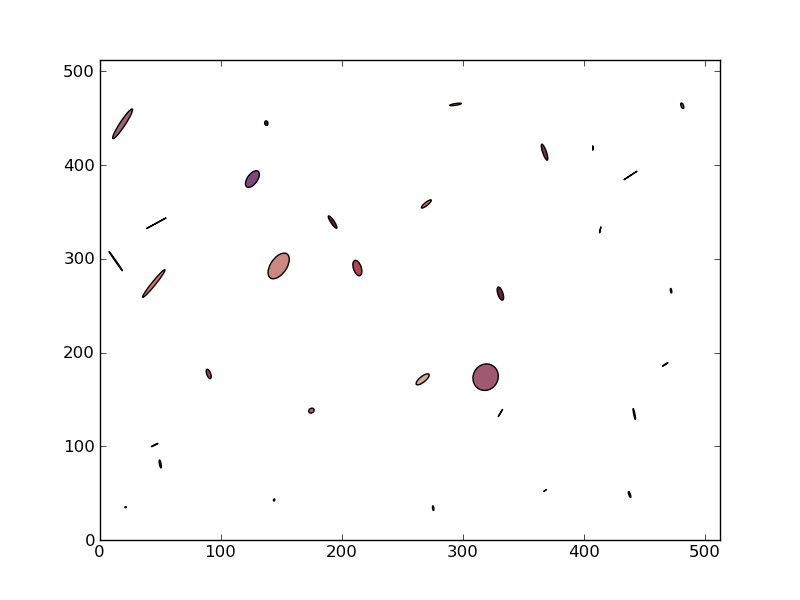}
(c)\includegraphics[bb=0 0 512 512,width=0.25\textwidth]{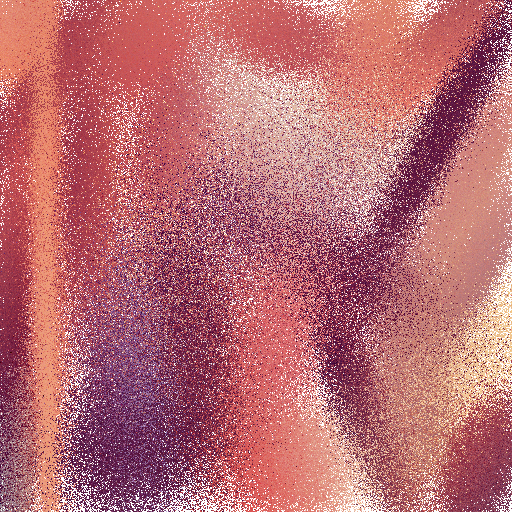}

\caption{A RGB color image (a) is interpreted as a 5D xyRGB point set on which a Gaussian mixture model (GMM) with $k=32$ components is trained (b).
Drawing many random variates from the generative GMM yields a sample image(c) that keeps low-frequency visual information.
\label{fig:GMM5D}}
\end{figure}

Figure~\ref{fig:GMM5D}(b) shows a GMM with $k=32$ components learned from a color image modeled as a 5D xyRGB point set (Figure~\ref{fig:GMM5D}(a)).
Since a GMM is a {\it generative model}, we can sample the GMM to create a ``sample image'' as shown in Figure~\ref{fig:GMM5D}(c).
Observe that low frequency information of the image is nicely modeled by GMMs.
Figure~\ref{fig:GMMhighD}(f) shows a GMM with $k=32$ components learned from a color image modeled as a high-dimensional point set.
Each $s\times s$ color image patch anchored at $(x,y)$ is modeled as a point in dimension $d=2+3s^2$.
GMM representations of images and videos~\cite{VideoGMM:2004} provide a compact feature representation that can be used in many applications, like  in information retrieval (IR) engines~\cite{Blobworld:2002}.

In this paper, we consider the general case of mixtures of distributions belonging the same exponential family~\cite{EMEF:1974}, like 
   Gaussian mixture models~\cite{GMM-CauchySchwarz-2011} (GMMs),  Rayleigh mixture models~\cite{RMM:2011} (RMMs),  Laplacian mixture models (LMMs)\cite{LaplacianMM-2007}, Bernoulli mixture models~\cite{BernoulliMM-2006} (BMMs), Multinomial Mixture models~\cite{MMM:2007} (MMMs), 
   Poisson Mixture Models (PMMs)~\cite{PMM:2006}, Weibull Mixture Models~\cite{WeiMM:2010} (WeiMMs), Wishart Mixture Models~\cite{WisMM:2012} (WisMM), etc. 
  
   \def\ttt{3cm}
\begin{figure} 
\centering
 \begin{tabular}{ccc}
\includegraphics[bb=0 0 512 512, width=\ttt]{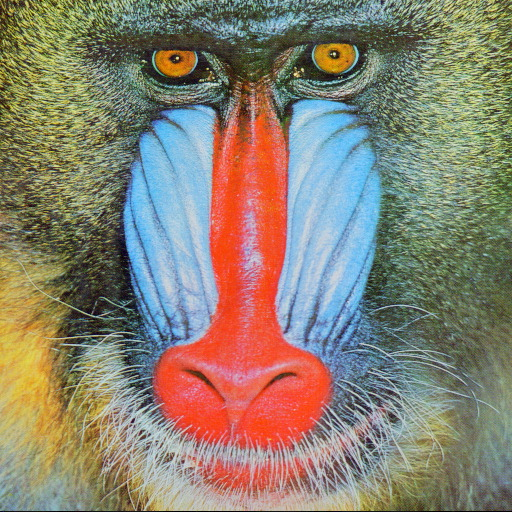} &
\includegraphics[bb=0 0 667 508, width=\ttt, height=\ttt]{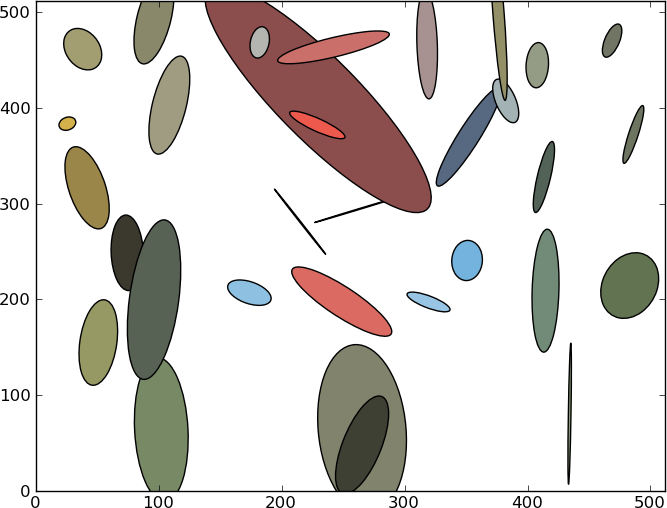}
&
\includegraphics[bb=0 0 512 512, width=\ttt]{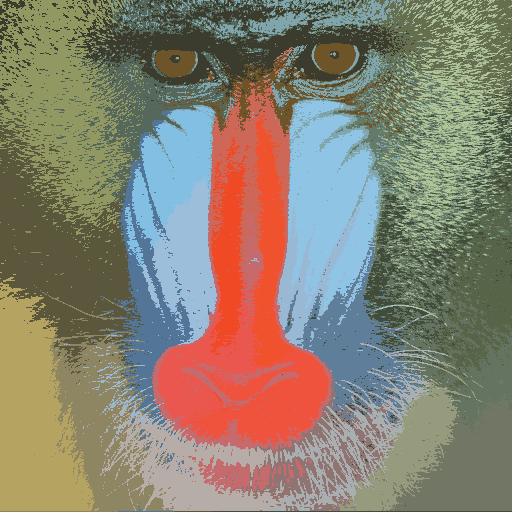}
\\
(a) & (b) & (c)\\
  \includegraphics[bb=0 0 512 512, width=\ttt]{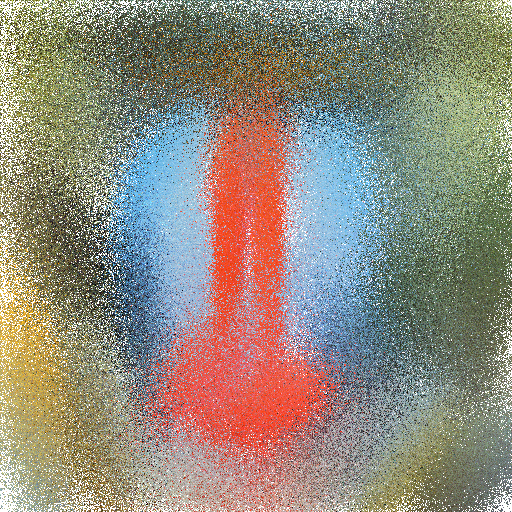}
&
\includegraphics[bb=0 0 512 512, width=\ttt]{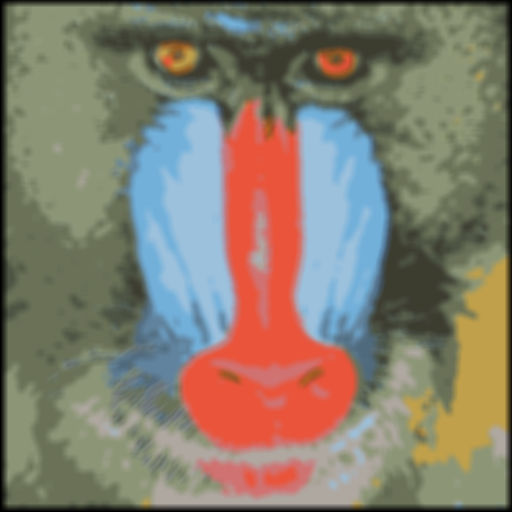} &
\includegraphics[bb=0 0 92 44, width=\ttt]{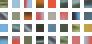}\\
\\
(d) & (e) & (f)
\end{tabular}  
\caption{Modeling a color image using a Gaussian mixture model (GMM):
(a) Image {\tt Baboon} source image, (b) a 5D $32$-GMM modeling depicted by its covariance ellipses,
(c) hard segmentation using the GMM, (d) sampling the 5D GMM,
(e) Mean colors ($8\times 8$ patches) for GMM with patch size $s=8$, (f) patch mean $\mu$ for $s=8$ patch size width.\label{fig:GMMhighD}}
\end{figure}

\subsection{Contributions and prior work}

Expectation-Maximization~\cite{em-1977} (EM) is a traditional algorithm for learning finite mixtures~\cite{FiniteMixtureModels:2000}.
Banerjee et al.~\cite{bregmankmeans-2005} proved that EM for mixture of exponential families amounts to perform equivalently a soft Bregman clustering.
Furthermore, this EM-Bregman soft clustering equivalence was extended to total Bregman soft clustering for curved exponential families~\cite{tBDPami:2012}.
Although mathematically convenient, we should remember that mixture data should be hard clustered as each observation should emanate from exactly one component.

It is well-known that $k$-means clustering technique can be interpreted as a limit case of EM for isotropic Gaussian mixtures~\cite{ClusteringMVN:2009}.
Kearns~et al.~\cite{HardSoftClustering:1997} casted further light on the hard/soft relationship  using an information-theoretic analysis of hard $k$-means and soft expectation-mazimization assignments in clustering.
Banerjee et al~\cite{BanerjeeMLEEF:2004} proved a mathematical equivalence between the estimation of maximum likelihood of exponential family mixtures (MLME, Maximum Likelihood Mixture Estimation) and a rate distortion problem for Bregman divergences.
Furthermore, Banerjee et al.~\cite{ClusteringvMF:2005} proposed the hardened expectation for the special case of von Mises-Fisher mixtures  (hard EM, Section 4.2 of~\cite{ClusteringvMF:2005}) for computational efficiency.

In this paper, we  build on the duality between Bregman divergences and exponential families~\cite{bregmankmeans-2005} to design $k$-MLE that iteratively (1) assigns data to mixture components, (2) update mixture parameters \`a la $k$-means and repeat step (1) until local convergence, (3) update weights and reiterate from (1) until local convergence (see Algorithm~\ref{algo:kmle}).
We prove that $k$-MLE maximizes monotonically the complete likelihood function.
We also discuss several initialization strategies and describe a probabilistic initialization $k$-MLE++ with guaranteed performance bounds.

The paper is organized as follows:
Section~\ref{sec:preliminaries} recall the basic notions of exponential families, Legendre transform, Bregman divergences, and demonstrate the duality  between Bregman divergences and exponential families to study the Maximum Likelihood Estimator (MLE).
Section~\ref{sec:kmle-theta} presents the framework of $k$-MLE for mixtures with prescribed weights, based on the Bregman-exponential family duality.
The generic $k$-MLE algorithm is described in Section~\ref{sec:fullkmle}, and Section~\ref{sec:speedup} discusses on proximity location data-structures to speed up the assignment step of the algorithm.
Section~\ref{sec:kmlepp} presents $k$-MLE++, a probabilistic initialization of $k$-MLE.
Finally, Section~\ref{sec:concl} concludes the paper and discusses on avenues for future research.

\section{Preliminaries}\label{sec:preliminaries}
 
\subsection{Exponential family}
   
An exponential family~\cite{BrownExpFam:1986} $E_F$ is a set of parametric probability distributions 
\begin{equation}
E_F=\{p_F(x;\theta)\ |\ \theta\in\Theta\}
\end{equation}
whose probability density\footnote{For sake of simplicity and brevity, we consider without loss of generality in the remainder continuous random variables on $\mathbb{R}^d$. We do not introduce the framework of probability measures nor Radon-Nikodym densities.} can be decomposed canonically as

\begin{equation}  \label{eq:cexpfam}
p_F(x;\theta) = e^{\inner{t(x)}{\theta}-F(\theta)+k(x)}
\end{equation}

where $t(x)$ denotes the sufficient statistics, $\theta$ the natural parameter, $F(\theta)$ the log-normalizer, and $k(x)$ a term related to an optional auxiliary carrier measure. $\inner{x}{y}$ denotes the inner product (i.e., $x^T y$ for vectors $\tr(X^T Y)$ for matrices, etc.). 
Let 
\begin{equation}
\Theta=\left\{\theta\ |\ \int p_F(x;\theta)\dx < \infty\right\}
\end{equation}
denotes the natural parameter space. 
The dimension $D$ of the natural parameter space is called the order of the family.
For the $d$-variate Gaussian distribution, the order is $D=d+\frac{d(d+1)}{2}=\frac{d(d+3)}{2}$.
It can be proved using the Cauchy-Schwarz inequality~\cite{BrownExpFam:1986} that the log-normalizer\footnote{Also called in the literature as the log-partition function, the cumulant function, or the log-Laplace function.} $F$ is a strictly convex and differentiable function on an open convex set $\Theta$.
The log-density of an exponential family is

\begin{equation}
l_{F}(x;\theta)= \inner{t(x)}{\theta}-F(\theta)+k(x)
\end{equation}

To build an exponential family, we need to choose a basic density measure on a support $\X$, a sufficient statistic $t(x)$, and an auxiliary carrier measure term $k(x)$. Taking the log-Laplace transform, we get

\begin{equation}
F(\theta) = \int_{x\in\mathbb{X}} e^{\inner{t(x)}{\theta}+k(x)} \dx,
\end{equation}
and define the natural parameter space as the $\theta$ values ensuring convergence of the integral.

In fact, many usual statistical distributions such as the Gaussian, Gamma, Beta, Dirichlet, Poisson, multinomial, Bernoulli, von Mises-Fisher, Wishart, Weibull are exponential families in disguise. 
In that case, we start from their probability density or mass function to retrieve the canonical decomposition of Eq.~\ref{eq:cexpfam}.
See~\cite{ef-flashcards-2009} for usual canonical decomposition examples of some distributions that includes a bijective conversion function $\theta(\lambda)$ for going from the usual $\lambda$-parameterization of the distribution to the $\theta$-parametrization.

Furthermore, exponential families can be parameterized canonically either using the natural coordinate system $\theta$, or by using the dual moment parameterization $\eta$ (also called mean value parameterization) arising from the Legendre transform (see Appendix~\ref{sec:mvn} for the case of Gaussians).

\subsection{Legendre duality and convex conjugates}

For a strictly convex and differentiable function $F:\thetaspace\rightarrow \mathbb{R}$, we define its convex conjugate by

\begin{equation}
F^*(\eta) = \sup_{\theta\in\thetaspace} \{ \underbrace{\inner{\eta}{\theta}-F(\theta)}_{l_F(\eta;\theta)} \}
\end{equation}

The maximum is obtained for $\eta=\nabla F(\theta)$ and is unique since $F$ is convex $\nabla^2_\theta l_F(\eta;\theta)  = -\nabla ^2 F(\theta) \prec 0$:

\begin{equation}
\nabla_\theta l_F(\eta;\theta) = \eta-\nabla F(\theta)=0  \Rightarrow \eta=\nabla F(\theta)
\end{equation}

Thus strictly convex and differentiable functions come in pairs $(F,F^*)$ with gradients being functional inverses of each other $\nabla F=(\nabla F^*)^{-1}$ and $\nabla F^*=(\nabla F)^{-1}$.
Legendre transform is an involution: ${(F^*)}^*=F$ for strictly convex and differentiable functions.
In order to compute $F^*$, we only need to find the functional inverse $(\nabla F)^{-1}$ of $\nabla F$ since 
 
\begin{equation}\label{eq:Fdual}
F^{*}(\eta) =  \inner{(\nabla F)^{-1}(\eta)}{\eta} - F((\nabla F)^{-1}(\eta)).
\end{equation}

However, this inversion may require numerical solving when no analytical expression of $\nabla F^{-1}$ is available.
See for example the gradient of the log-normalizer of the Gamma distribution~\cite{ef-flashcards-2009}, the Dirichlet or von Mises-Fisher distributions~\cite{ClusteringvMF:2005}.

\subsection{Bregman divergence}

A Bregman divergence $B_F$ is defined for a strictly convex and differentiable generator $F$ as

\begin{equation}
B_F(\theta_1 : \theta_2) = F(\theta_1)-F(\theta_2)-\inner{\theta_1-\theta_2}{\nabla F(\theta_2)}.
\end{equation}
 
The Kullback-Leibler divergence (relative entropy) between two members $p_1=p_F(x;\theta_1)$ and $p_2=p_F(x;\theta_2)$ of the same exponential family amounts to compute a Bregman divergence on the corresponding swapped natural parameters:

\begin{eqnarray}
\KL(p_1:p_2) &=& \int_{x\in\mathbb{X}} p_1(x)\log\frac{p_1(x)}{p_2(x)}\dx,\\
&=& B_F(\theta_2:\theta_1),\\
 &=& F(\theta_2)-F(\theta_1)-\inner{\theta_2-\theta_1}{\nabla F(\theta_1)}
\end{eqnarray}

The proof follows from the fact that $E[t(X)]=\int_{x\in\mathbb{X}} t(x)p_F(x;\theta)\dx=\nabla F(\theta)$~\cite{CrossEntropy:2010}.
Using Legendre transform, we further have the following equivalences of the relative entropy: 

\begin{eqnarray}
B_F(\theta_2:\theta_1) &=& B_{F*}(\eta_1 :\eta_2),\\
&=& \underbrace{ F(\theta_2) + F^*(\eta_1) -\inner{\theta_2}{\eta_1}}_{C_F(\theta_2 : \eta_1)=C_{F^*}(\eta_1:\theta_2)} \label{eq:cd},
\end{eqnarray}
where $\eta=\nabla F(\theta)$ is the dual moment parameter (and $\theta=\nabla F^*(\eta)$).
Information geometry~\cite{informationgeometry-2000} often considers the canonical divergence $C_F$ of Eq.~\ref{eq:cd} that uses the mixed coordinate systems $\theta/\eta$, while computational geometry~\cite{bvd-2010} tends to consider dual Bregman divergences, $B_F$ or $B_{F^*}$, and visualize structures in one of those two canonical coordinate systems.
Those canonical coordinate systems are dually orthogonal since $\nabla^2 F(\theta) \nabla^2 F^*(\eta)=I$, the identity matrix.

\subsection{Maximum Likelihood Estimator (MLE)}

For exponential family mixtures with a single component $M\sim E_F(\theta_1)$ ($k=1$, $w_1=1$), we easily estimate the parameter $\theta_1$.
Given $n$ independent and identically distributed observations $x_1, ..., x_n$, the Maximum Likelihood Estimator (MLE) is  maximizing the likelihood function:

\begin{eqnarray}
\hat\theta &=& \argmax_{\theta\in\Theta}  L(\theta; x_1, ..., x_n), \\
& =& \argmax_{\theta\in\Theta} \prod_{i=1}^n p_F(x_i;\theta),\\
&=& \argmax_{\theta\in\Theta} e^{\sum_{i=1}^n \inner{t(x_i)}{\theta}-F(\theta)+k(x_i)}
\end{eqnarray}

For exponential families, the MLE reports a unique maximum since the Hessian of $F$ is positive definite ($X\sim E_F(\theta) \Rightarrow \nabla^2 F = \var[t(X)]\succ 0$):
\begin{equation}
\nabla F(\hat\theta) = \frac{1}{n} \sum_{i=1}^n t(x_i)
\end{equation}

The MLE is consistent and efficient with asymptotic normal distribution: 

\begin{equation}
\hat\theta \sim N \left(\theta,\frac{1}{n} I_F^{-1}(\theta)\right),
\end{equation}
where $I_F$ denotes the Fisher information matrix:

\begin{equation}
I_F(\theta) = \mathrm{var}[t(X)] = \nabla ^2 F(\theta) =(\nabla^2 G(\eta))^{-1}
\end{equation}
(This proves the convexity of $F$ since the covariance matrix is necessarily positive definite.)
Note that the MLE may be biased  (for example, normal distributions).

By using the Legendre transform, the  log-density of an exponential family can be interpreted as a Bregman divergence~\cite{bregmankmeans-2005}:
\begin{equation}
\log  p_F(x;\theta)  =  -B_{F^*}(t(x) : \eta) + F^*(t(x)) + k(x) 
\end{equation}

Table~\ref{tab:duality} reports some illustrating examples of the Bregman divergence $\leftrightarrow$ exponential family duality.
\begin{table}
\centering
\begin{tabular}{|ccc|}\hline
  Exponential Family & $\Leftrightarrow$ & Dual Bregman divergence \\ 
    $p_F(x|\theta)$ &  &$B_{F^*}$\\ \hline
  Spherical Gaussian & $\Leftrightarrow$ & Squared Euclidean divergence\\
   Multinomial & $\Leftrightarrow$ & Kullback-Leibler divergence\\
   Poisson &  $\Leftrightarrow$  & $I$-divergence\\
 Geometric & $\Leftrightarrow$ & Itakura-Saito divergence\\
  Wishart & $\Leftrightarrow$ & log-det/Burg matrix divergence\\ \hline  
\end{tabular}
\caption{Some examples illustrating the duality between exponential families and Bregman divergences.\label{tab:duality}}
\end{table}
Let us use the Bregman divergence-exponential family duality to prove that 
\begin{equation}
\hat\theta=\arg\max_{\theta\in\Theta} \prod_{i=1}^n p_F(x_i;\theta)=\nabla F^{-1}\left(\sum_{i=1}^n t(x_i)\right).
\end{equation}
 
Maximizing the average log-likelihood $\bar l=\frac{1}{n} \log L$, we have:
 
\begin{eqnarray}
&\max_{\theta\in\thetaspace} & \bar l(\theta;x_1, ..., x_n)=\frac{1}{n} \sum_{i=1}^n (\inner{t(x_i)}{\theta}-F(\theta)+k(x_i)) \\
&\max_{\theta\in\thetaspace} &  \frac{1}{n} \sum_{i=1}^n -B_{F^*}(t(x_i):\eta)+F^*(t(x_i)) + k(x_i)\\
&\equiv \min_{\eta\in\etaspace} & \frac{1}{n} \sum_{i=1}^n B_{F^*}(t(x_i):\eta)
\end{eqnarray}

Since right-sided Bregman centroids defined as the minimum average divergence minimizers coincide always with the center of mass~\cite{bregmankmeans-2005} (independent of the generator $F$), it follows that 
\begin{equation}
\hat\eta=\frac{1}{n}\sum_{i=1}^n t(x_i)=\nabla F(\hat\theta).
\end{equation}
It follows that $\hat\eta=(\nabla F)^{-1}(\frac{1}{n}\sum_{i=1}^n t(x_i))$.
 
In information geometry~\cite{informationgeometry-2000}, the point $\hat P$ with $\eta$-coordinate $\hat\eta$ (and $\theta$-coordinate $\nabla F^{-1}(\hat\eta)=\hat\theta$) is called the {\em observed} point. 
The best average log-likelihood reached by the MLE  at $\hat\eta$ is 

\begin{eqnarray}
l(\hat\theta; x_1,..., x_n) &=& \frac{1}{n} \sum_{i=1}^n ( -B_{F^*}(t(x_i):\hat\eta)+F^*(t(x_i)) + k(x_i) ), \\
&=& \frac{1}{n} \sum_{i=1}^n (-F^*(t(x_i))+F^*(\hat\eta)+\inner{t(x_i)-\hat\eta}{\nabla F^*(\hat\eta)}+F^*(t(x_i)) + k(x_i)),\\
&=& F^*(\hat\eta)+\frac{1}{n}\sum_{i=1}^n k(x_i)+\Inner{\underbrace{\frac{1}{n}\sum_{i=1}^n t(x_i)-\hat\eta}_{0}}{\hat\theta},\\
&= &  F^*(\hat\eta) +\frac{1}{n}\sum_{i=1}^n k(x_i).
\end{eqnarray} 

The Shannon entropy $H_F(\theta)$ of $p_F(x;\theta)$ is $H_F(\theta)=-F^*(\eta)-\int k(x)p_F(x;\theta)\dx$~\cite{CrossEntropy:2010}.
Thus the maximal likelihood is related to the minimum  entropy (i.e., reducing the uncertainty) of the empirical distribution. 

Another proof follows from the Appendix~\ref{sec:kmeans} where it is recalled that the Bregman information~\cite{bregmankmeans-2005} (minimum of average right-centered Bregman divergence) obtained for the center of mass is a Jensen diversity index.
Thus we have 

\begin{eqnarray}
\bar l &=& -J_{F^*}(\sum_{i=1}^n t(x_i))+\frac{1}{n} \sum_{i=1}^n F^*(t(x_i)) + \frac{1}{n} \sum_{i=1}^n k(x_i),\\
&=& - \left(\sum_{i=1}^n F^*(t(x_i)) -F^*(\hat\eta)\right) +\frac{1}{n} \sum_{i=1}^n F^*(t(x_i)) + \frac{1}{n} \sum_{i=1}^n k(x_i),\\
&=& F^*(\hat\eta)+  \frac{1}{n} \sum_{i=1}^n k(x_i)
\end{eqnarray}

Appendix~\ref{sec:mvn} reports the dual canonical parameterizations of the multivariate Gaussian distribution family.

\section{$k$-MLE: Learning mixtures with given prescribed weights}\label{sec:kmle-theta}

Let $\X=\{x_1, ..., x_n\}$ be a sample set of independently and identically distributed observations from a finite mixture $m(x|w,\theta)$ with $k$ components.
The joint probability distribution of the observed observations $x_i$'s with the missing component labels $z_i$'s  is

\begin{equation}
p(x_1, z_1, ..., x_n, z_n | w,\theta) = \prod_{i=1}^n p(z_i | w) p(x_i | z_i, \theta)
\end{equation}

To optimize the joint distribution, we could test (theoretically) all the $k^n$ labels, and choose the best assignment.
This is not tractable in practice since it is exponential in $n$ for $k>1$.
Since we do not observe the latent variables $z_1, ..., z_n$, we marginalize the hidden variables to get

\begin{equation}
p(x_1, ..., x_n | w,\theta) = \prod_{i=1}^n \sum_{j=1}^k p(z_i=j|w) p(x_i|z_i=j,\theta_j)
\end{equation}

The average log-likelihood function is 

\begin{eqnarray}
\bar l(x_1, ..., x_n | w,\theta) &=& \frac{1}{n} \log p(x_1, ..., x_n | w,\theta),\\
&=& \frac{1}{n} \sum_{i=1}^n \log \sum_{j=1}^k p(z_i=j|w) p(x_i|z_i=j,\theta_j).
\end{eqnarray}

Let $\delta_j(z_i)=1$ if and only if $x_i$ has been sampled from the $j$th component, and $0$ otherwise.
We have the complete average log-likelihood that is mathematically rewritten as

\begin{eqnarray}
\bar l(x_1, z_1, ..., x_n, z_n | w,\theta) &=& \frac{1}{n} \sum_{i=1}^n \log \prod_{j=1}^k (w_j p_F(x_i|\theta_j))^{\delta_j(z_i)} \\
&=& \frac{1}{n} \sum_{i=1}^n \sum_{j=1}^k \delta_j(z_i) (\log p_F(x_i|\theta_j) + \log w_j)
\end{eqnarray}

Using the bijection between exponential families and dual Bregman divergences~\cite{bregmankmeans-2005}, we have the mathematical equivalence $\log p_F(x|\theta_j) = -B_{F^*}(t(x):\eta_j)+F^*(t(x))+k(x)$, where $\eta_j=\nabla F(\theta_j)$ is the moment parameterization of the $j$-th component  exponential family distribution.
It follows that the complete  average log-likelihood function is written as

\begin{eqnarray}
\bar l(x_1, ..., x_n | w,\theta) &=& \frac{1}{n} \sum_{i=1}^n \sum_{j=1}^k \delta_j(z_i) (-B_{F^*}(t(x_i):\eta_j)+F^*(t(x_i))+k(x_i)+ \log w_j)\\
&=&  \left( \frac{1}{n} \sum_{i=1}^n \sum_{j=1}^k \delta_j(z_i) (-B_{F^*}(t(x_i):\eta_j)+ \log w_j) \right) +  \frac{1}{n} \sum_{i=1}^n   F^*(t(x_i))+k(x_i). \label{eq:ll}
\end{eqnarray}

By removing the constant terms $ \frac{1}{n}\sum_{i=1}^n  (F^*(t(x_i))+k(x_i))$ independent of the mixture moment parameters (the $\eta$'s), 
maximizing the complete average log-likelihood amounts to equivalently minimize the following loss function:

\begin{eqnarray}
\bar l' &=& \frac{1}{n} \sum_{i=1}^n \sum_{j=1}^k \delta_j(z_i) (B_{F^*}(t(x_i):\eta_j)-\log w_j),\\
&= &  \frac{1}{n} \sum_{i=1}^n \min_{j=1}^k  (B_{F^*}(y_i:\eta_j)-\log w_j), \label{eq:dualkmeans}\\
&=& \kmeans_{F^*,\log w}(\Y : H),
\end{eqnarray}
where $\Y=\{y_1=t(x_1), ..., y_n=t(x_n( \}$ and $H=\{\eta_1, ..., \eta_k\}$. 

\begin{remark}
This is the argmin of Eq.~\ref{eq:dualkmeans} that gives the hidden component labels for the $x_i$'s.
\end{remark}

\begin{remark}
Observe that since $\forall i\in\{1, ...,k\}, -\log w_i \geq 0$ (since $w_i\leq 1$),  we have the following additive dual Bregman divergence $B_{F^*}(y_i:\eta_j)-\log w_j>0$ per cluster.
Depending on the weights (e.g., $w\rightarrow 0$), we may have some empty clusters.
In that case, the weight of a cluster is set to zero (and the component parameter is set to $\emptyset$ by convention).
Note that it makes sense to consider $(\leq k)$-means instead of $k$-means in the sense that we would rather like to upper bound the maximum complexity of the model rather than precisely fixing it.
\end{remark}

Eq.~\ref{eq:dualkmeans} is precisely the loss function of a per-cluster additive Bregman $k$-means (see the appendix~\ref{sec:kmeans}) defined for the Legendre convex conjugate $F^*$ of the log-normalizer $F$ of the exponential family for the sufficient statistic points $\mathcal{Y} = \{y_i=t(x_i)\}_{i=1}^n$.
It follows that {\em any} Bregman $k$-means heuristic decreases monotonically  the loss function  and reaches a local minimum (corresponding to a local maximum for the equivalent complete likelihood function). 
We can either use the batched  Bregman Lloyd's $k$-means~\cite{bregmankmeans-2005}, the Bregman Hartigan and Wong's greedy cluster swap heuristic~\cite{KmeansHartiganWong-1979,HartiganKmeans:2010}, or the Kanungo et al.~\cite{KanungoKmeans-2004} $(9+\epsilon)$-approximation global swap approximation algorithm.

\begin{remark}
The likelihood function $L$ is equal to $e^{n\bar l}$. 
The average likelihood function $\bar L$ is defined by taking the geometric mean $\bar L=L^{\frac{1}{n}}$.
\end{remark}

The following section shows how to update the weights once the local convergence of the assignment-$\eta$ of the $k$-MLE loop has been reached.

\section{General $k$-MLE including mixture weight updates}\label{sec:fullkmle}

When $k$-MLE with prescribed weights reaches a local minimum (see Eq.~\ref{eq:ll} and Eq.~\ref{eq:dualkmeans} and the appendix~\ref{sec:kmeans}), the current loss function is equal to

\begin{eqnarray}
\bar l=\underbrace{\frac{1}{n} \sum_{i=1}^n \sum_{j=1}^k \delta_j(z_i) (B_{F^*}(t(x_i):\eta_j) - \log w_j)}_{\mbox{Minimized by  additive  Bregman $k$-means, see Appendix}} &-& \left( \frac{1}{n} \sum_{i=1}^n   F^*(t(x_i))+k(x_i) \right),\\
\bar l=\sum_{j=1}^k \alpha_j J_{F^*}(\C_j)-\alpha_j\log w_j  &-& \left(\frac{1}{n}\sum_{i=1}^n   F^*(t(x_i))+k(x_i)\right),\label{eq:llw}
\end{eqnarray}
where $\alpha_i=\frac{|\C_i|}{n}$ denotes the proportion of points assigned to the $i$-th cluster $\C_i$, and $\alpha_i J_{F^*}(\C_i)$ is the weighted Jensen diversity divergence of the cluster.
In order to further minimize the average complete likelihood of Eq.~\ref{eq:llw}, we update the mixture weights $w_i$'s by minimizing the criterion:

\begin{eqnarray}
&&\min_{w\in\Delta_k} \sum_{j=1}^k -\alpha_j\log w_j \\
&=& \min_{w\in\Delta_k} H^{\times}(\alpha:w),
\end{eqnarray}
where $H^{\times}(p:q)=-\sum_{i=1}^k p_i\log q_i$ denotes the Shannon cross-entropy, and $\Delta_k$ the $(k-1)$-dimensional probability simplex.
The cross-entropy $H^\times(p:q)$ is minimized for $p=q$, and yields $H^\times(p,p)=H(p)=-\sum_{i=1}^k p_i \log p_i$, the Shannon entropy.
Thus we update the weights 
by taking the relative proportion of points falling into the clusters: 

\begin{equation}
\forall i\in\{1, ...,k\}, w_i\leftarrow \alpha_i.
\end{equation}

After updated the weights, the average complete log-likelihood is

\begin{equation}
\bar l= \sum_{i=1}^k w_i J_{F^*}(\C_i) + H(w) - \left( \frac{1}{n}\sum_{i=1}^n   F^*(t(x_i))+k(x_i) \right).
\end{equation}

We summarize the $k$-MLE algorithm in the boxed Algorithm~\ref{algo:kmle}.

\begin{algo}
\caption{Generic $k$-MLE for learning an exponential family mixture model.\label{algo:kmle}}

\underline{Input}:\\
\begin{tabular}{lll}
$\X$ &:&  a set of $n$ identically and independently distributed observations: $\X=\{x_1, ..., x_n\}$\\
$F$ &:& log-normalizer of the exponential family, characterizing $E_F$ \\
$\nabla F$ &:& gradient of $F$ for moment $\eta$-parameterization: $\eta=\nabla F(\theta)$\\
$\nabla F^{-1}$ &:& functional inverse of the gradient of $F$ for $\theta$-parameterization: $\theta=\nabla F^{-1}(\eta)$\\
$t(x)$ &:& the sufficient statistic of the exponential family\\
$k$ &:& number of clusters\\
\end{tabular}

\begin{itemize}
\item 0. {\bf Initialization}: $\forall i\in\{1, ..., k\},$ let  $w_i=\frac{1}{k}$ and $\eta_i=t(x_i)$\\
(Proper initialization is further discussed later on).

\item 1. {\bf Assignment}: $\forall i\in\{1, ...,n\}, z_i=\argmin_{j=1}^k B_{F^*}(t(x_i):\eta_j)-\log w_j$.\\
Let $\forall i\in\{1, ..., k\}\ \C_i=\{x_j | z_j=i\}$ be the cluster partition: $\X=\cup_{i=1}^k \C_i$.\\
(some clusters may become empty depending on the weight distribution)

\item 2. {\bf Update the $\eta$-parameters}: 
$\forall i\in\{1, ..., k\}, \eta_i=\frac{1}{|\C_i|}\sum_{x\in\C_i} t(x)$.\\
(By convention, $\eta_i=\emptyset$ if $|\C_i|=0$)
{\bf Goto step~1} unless local convergence of the complete likelihood is reached.

\item 3. {\bf Update the mixture weights}: 
$\forall i\in\{1, ..., k\}, w_i=\frac{1}{n}|\C_i|$.\\
{\bf Goto step~1} unless local convergence of the complete likelihood is reached.
\end{itemize}

\underline{Output}: An exponential family mixture model $m(x)$ (EFMM) parameterized in the natural coordinate system: $\forall i\in\{1,...,k\}, \theta_i=(\nabla F)^{-1}(\eta_i)=\nabla F^*(\eta_i)$:
$$
m(x) = \sum_{i=1}^k w_i p_F(x|\theta_i)
$$

\end{algo}
 
\begin{remark}
Note that we can also do after the assignment step of data to clusters both (i) the mixture $\eta$-parameter  update and (ii) 
the mixture $w$-weight  update consecutively in a single iteration of the  $k$-MLE loop.
This corresponds to the Bregman hard expectation-maximization (Bregman Hard EM) algorithm  described in boxed Algorithm~\ref{algo:hardEM}. 
This Hard EM algorithm is straightforwardly implemented in legacy source codes by hardening the weight membership in the E-step of the EM.
Hard EM was shown computationally efficient when learning mixtures of von-Mises Fisher (vMF) distributions~\cite{ClusteringvMF:2005}.
Indeed,  the log-normalizer $F$ (used when computing densities) of vMF distributions requires to compute a modified Bessel function of the first kind~\cite{vMF:2011}, that is only invertible approximately using numerical schemes.  
\end{remark}

\begin{algo}
\caption{Hard EM for learning an exponential family mixture model.\label{algo:hardEM}} 

\begin{itemize}
\item 0. {\bf Initialization}: $\forall i\in\{1, ..., k\},$ let  $w_i=\frac{1}{k}$ and $\eta_i=t(x_i)$\\
(Proper initialization is further discussed  later on).

\item 1. {\bf Assignment}: $\forall i\in\{1, ...,n\}, z_i=\argmin_{j=1}^k B_{F^*}(t(x_i):\eta_j)-\log w_j$.\\
Let $ \forall i\in\{1, ..., k\}\ \C_i=\{x_j | z_j=i\}$ be the cluster partition: $\X=\cup_{i=1}^k \C_i$.

\item 2. {\bf Update the $\eta$-parameters}: 
$\forall i\in\{1, ..., k\}, \eta_i=\frac{1}{|\C_i|}\sum_{x\in\C_i} t(x)$.

\item 3. {\bf Update the mixture weights}: 
$\forall i\in\{1, ..., k\}, w_i=\frac{|\C_i|}{n}$.

\item {\bf Goto step~1} unless local convergence of the complete likelihood is reached.
\end{itemize}


\end{algo}

We can also sparsify EM by truncating to the first $D$ entries on each row (thus, we obtain a well-defined centroid per cluster for non-degenerate input). This is related to the sparse EM proposed in~\cite{SparseEM:1998}.
Degeneraties of the EM GMM is identified and discussed in~\cite{EMConvergence:2003}.
Asymptotic convergence rate of the EM GMM is analyzed in~\cite{EM-GMM-convergence:2001}.

There are many ways to initialize $k$-means~\cite{kmeansInit:1999}. 
Initialization shall be discussed in Section~\ref{sec:kmlepp}.

\section{Speeding up $k$-MLE and Hard EM using Bregman NN queries}\label{sec:speedup}

The proximity cells $\{\V_1, ..., \V_k\}$ induced by the cluster centers $\C=\{c_1, ..., c_k\}$ (in the $\eta$-coordinate system) are defined by:

\begin{equation}
\V_j  = \left\{ x\in\mathbb{X} \ |\ B_{F^*}(t(x):\eta_j)-\log w_j \leq B_{F^*}(t(x):\eta_l)-\log w_l, \forall l\in\{1, ...,k\}\backslash\{j\} \right\} 
\end{equation}
partitions the support $\mathbb{X}$ into a Voronoi diagram. 
It is precisely equivalent to the intersection of a Bregman Voronoi diagram for the dual log-normalizer $F^*$ with additive weights~\cite{bvd-2010} on the expectation parameter space $\etaspace=\{\eta=\nabla F(\theta)\ | \ \theta\in\thetaspace \}$ with the hypersurface\footnote{Note that there is only one global minimum for the distance $B_{F^*}(y:\eta)$ with $y\in\mathbb{T}$.} $\mathbb{T}=\{t(x)\ |\ x\in\mathbb{X}\}$. 
For the case of Gaussian mixtures, the log-density of the joint distribution $w_i p_F(x;\mu_i,\Sigma_i)$ induces a partition of the space into an
anisotropic weighted Voronoi diagram~\cite{AnisotropicVoronoiDiagram:2003}.
This is easily understood by taking {\em minus the log-density} of the Gaussian distribution (see Eq.~\ref{eq:mvn}):

\begin{equation} 
-\log p(x;\mu_i,\Sigma_i)= \frac{1}{2} D_{\Sigma_i^{-1}}(x-\mu_i,x-\mu_i) + \frac{1}{2}\log |\Sigma_i| +\frac{d}{2}\log 2\pi,
\end{equation}
with $M_Q$ the squared Mahalanobis distance $M_{Q}(x,y)=(x-y)^T Q (x-y)$.
This is an additively weighted Bregman divergence with mass $m_i=\frac{1}{2}\log |\Sigma_i| +\frac{d}{2}\log 2\pi$ and  generator $F_i(x)=\inner{x}{\Sigma_i^{-1} x}$, the precision matrix (see the Appendix).
Figure~\ref{fig:voronoi} displays the anisotropic Voronoi diagram~\cite{AnisotropicVoronoiDiagram:2003} of a 5D xyRGB GMM restricted to the xy plane.
We color each pixel with the mean color of the anisotropic Voronoi cell it belongs to.

\begin{figure}
\centering
(a)\includegraphics[bb=0 0 512 512,width=0.45\textwidth]{Figure/baboon.png}
(b)\includegraphics[bb=0 0 512 512,width=0.45\textwidth]{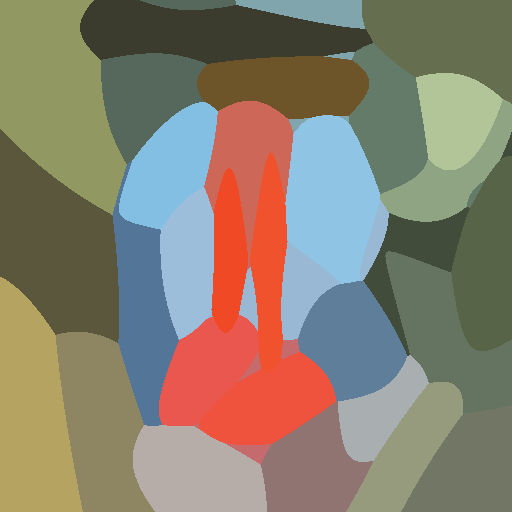}
\caption{From the source color image (a), we buid a 5D GMM with $k=32$ components, and color each pixel with the mean color of the anisotropic Voronoi cell it belongs to.\label{fig:voronoi}}
\end{figure}

When the order of the exponential family (i.e., number of parameters) is small (say, $D\leq 3$), we can compute explicitly this additively weighted Bregman Voronoi diagrams in the moment parameter space $\etaspace$, and use proximity location data-structures designed for geometric partitions bounded by planar walls.
Otherwise, we  speed up the assignment step of $k$-MLE/Hard EM by using proximity location data-structures such as Bregman ball trees~\cite{2009-W-BregmanTrees-EuroCG} or Bregman vantage point trees~\cite{2009-BregmanVantagePointTree-IEEE}.
See also~\cite{BregmanSearch:2011}.

Besides Lloyd's batched $k$-means heuristic~\cite{LLoyd:1982,MacQueen:1967,Forgy-1965}, we can also implement other $k$-means heuristic like the greedy Hartigan and Wong's  swap~\cite{KmeansHartiganWong-1979,HartiganKmeans:2010} in $k$-MLE that selects a point and optimally reassign it, or Kanungo et al.~\cite{KanungoKmeans-2004} global swap optimization, etc.

\begin{remark}
The MLE equation $\hat\eta=\nabla F(\hat\theta)=\frac{1}{n}\sum_{i=1}^n t(x_i)$ may yield a transcendental equation.
That is, when $(\nabla F)^{-1}$ is not  available analytically (e.g., von Mises-Fisher family~\cite{ClusteringvMF:2005}), the convex conjutate $F^*$ needs to be approximated by computing numerically the reciprocal gradient $\nabla F^{-1}$ (see Eq.~\ref{eq:Fdual}). 
Sra~\cite{vMF:2011} focuses on solving efficiently the MLE equation\footnote{See also, software R package {\tt movMF}} for the von Mises-Fisher distributions.
\end{remark}

\section{Initializing $k$-MLE using $k$-MLE++}\label{sec:kmlepp}

To complete the description of $k$-MLE of boxed Algorithm~1, it remains the problem to properly initializing $k$-MLE (step $0$). 
One way to perform this initialization is to compute the global MLE parameter for the full set $\X$:

\begin{equation}
\hat\eta=\nabla F^{-1}\left(\frac{1}{n}\sum_{i=1}^n t(x_i)\right),
\end{equation}
and then consider the {\em restricted exponential family} of order $d\leq D$ with {\em restricted sufficient statistic} the first $d$ components of full family statistic $(t_1(x), ..., t_d(x))$. We initialize the $i$-th cluster with $\eta_i^{(0)}=(t_1(x_i), ..., t_d(x_i), \hat\eta_{d+1}, ..., \hat\eta_D)$.
For the case of multivariate Gaussians with $D=\frac{d(d+3)}{2}$, this amounts to compute the covariance matrix $\hat\Sigma$ of the full  set and then set the translation parameter to $x_i$: $\eta_i^{(0)}=(x_i,-\frac{1}{2}(\hat\Sigma+x_ix_i^T))$ (see appendix~\ref{sec:mvn}). 
This initialization is a heuristic with {\it no guaranteed performance} on the initial average complete log-likelihood $\bar l$ compared to the best one $\bar l^*$. 
Note that when $D=d$ (e.g., Poisson, Weibull, Rayleigh, isotropic Gaussian, etc.), we need to have distinct initializations so that instead of taking the global MLE, we rather split the data set into $k$ groups of size $\frac{n}{k}$, and take the MLE of each group for initialization.
A good geometric split is given by using a Voronoi partition diagram as follows:
We run Bregman $k$-means on $\Y$ for the dual convex conjugate $F^*$ and set the mixture parameters as the MLEs of clusters and the weights as the relative proportion of data in clusters. 
This corroborates an experimental observation by Banerjee et al.~\cite{bregmankmeans-2005} that observes that clustering works experimentally best if we choose the dual Bregman divergence associated with the exponential family mixture sample set.

Let us further use   the dual Bregman $k$-means interpretation of EM to perform this initialization efficiently.
Assume uniform weighting of the mixtures. 
That is, $\forall i\in\{1, ...,k\}, w_i=\frac{1}{k}$.

Maximizing the average complete log-likelihood  amounts to minimize (see Eq.~\ref{eq:dualkmeans}):

\begin{equation}
\bar l'' = \frac{1}{n} \sum_{i=1}^k \min_{j=1}^k  B_{F^*}(y_i=t(x_i):\eta_j).
\end{equation}

The likelihood function $L(x_1, ..., x_n |\theta,w)$ is

\begin{equation}
L = e^{-n\kmeans_{F^*}(\C)+n\log k+\sum_{i=1}^n (F^*(x_i)+k(x_i))}.
\end{equation} 

Thus for uniform mixture weights, the ratio between two different $k$-means optimization with respective cluster centers $\C$ and $\C'$ is:

\begin{equation}
\frac{L}{L'} = e^{-n (\kmeans_{F^*}(\C)-\kmeans_{F^*}(\C')) }
\end{equation}

We can use the standard Bregman $k$-means++ initialization~\cite{BregmanClustering-2010} on the convex conjugate $F^*$ that gives probabilistically a guaranteed $O(\mu^{-2} \log k)$ performance, where $\mu$ is a constant factor to be explained below. 
The Bregman $k$-means++ algorithm is recalled in boxed Algorithm~\ref{algo:bregkmeanspp}.

\begin{algo}
\caption{Bregman $k$-means++: probabilistically guarantees a good initialization.\label{algo:bregkmeanspp}}
 
\begin{itemize}

\item Choose first seed $\C=\{y_l\}$, for $l$ uniformly random in $\{1, ..., n\}$.

\item For $i\leftarrow 2$ to $k$

\begin{itemize}
\item Choose $c_i\in \{y_1, ..., y_n\}$ with probability 

$$
p_i = \frac{B_F(c_i:\C)}{\sum_{i=1}^n B_F(y_i:\C)}  = \frac{B_F(\Y:\C)}{\kmeans_F(\Y:\C)},
$$
where $B_F(c:\C)=\min_{p\in\C} B_F(c:p)$.

\item Add selected seed to the initialization seed set: $\C\leftarrow \C\cup \{c_i\}$, and reiterate until $|\C|=k$.

\end{itemize}
\end{itemize}

\end{algo}

Let $\kmeans_F^{*}$ denote the optimal Bregman $k$-means average loss function for generator $F$.
Bregman $k$-means++~\cite{BregmanClustering-2010} described in Algorithm~\ref{algo:bregkmeanspp} ensures that

\begin{equation}
{\kmeans_F}^*(\Y : \C) \leq \kmeans_F(\Y : \C) \leq \frac{8}{\mu^2} (2+\log k) {\kmeans_F}^*(\Y : \C)
\end{equation}

The factor $\mu$ in the upper bound is related to  the notion of $\mu$-similarity that we now concisely explain.
Observe that the squared Mahalanobis distance $M_Q(p,q) = (p-q)^T Q (p-q)$ satisfies the double triangle inequality:

\begin{equation}
M_Q(p,q) \leq 2 (M_Q(p,r)  + M_Q(r,q) ).
\end{equation}

A Bregman divergence is said to have the $\mu$-similarity on a domain $\Y$ if there exists a  positive definite matrix $Q\succ 0$ on $\Y=\mathrm{conv}(y_1, ..., y_n)$ and a real $0<\mu\leq 1$ such that

\begin{equation}
\mu M_Q(p,q) \leq B_F(p:q) \leq M_Q(p,q)
\end{equation}

Since a Bregman divergence can also be interpreted as the remainder of a Taylor expansion using the Lagrange error term:

\begin{equation}
B_F(p:q) = (p-q)^T \frac{\nabla ^2 F(\epsilon_{pq})}{2} (p-q),
\end{equation}
with $\epsilon_{pq}$ being a point on the line segment $[pq]$.
It follows that by considering the Hessian $\nabla^2 F$ on a compact subset $\Y=\mathrm{conv}(y_1, ..., y_n)$, we get a bound~\cite{MixedBregmanClustering:2008} for $\mu$ as follows:

\begin{equation}
\mu=\min_{p,q\in\Y} \frac{\min_{y\in\Y} (p-q)^T \nabla^2 F(y) (p-q)}{\max_{y\in\Y} (p-q)^T \nabla^2 F(y) (p-q)}.
\end{equation}

By considering a hyperrectangle bounding the convex hull $\Y=\mathrm{conv}(y_1, ..., y_n)$, it is usually easy to compute bounds for $\mu$.
See~\cite{BregmanClustering-2010} for some examples.

The notion of $\mu$-similarity also allows one to design fast proximity queries~\cite{BregmanSearch:2011} based on the following two properties:

\begin{description}

\item[Approximately symmetric.]

\begin{equation}
B_F(p:q) \leq \frac{1}{\mu} B_F(q,p)
\end{equation}

\item[Deficient triangle inequality.]
\begin{equation}
B_F(p:q) \leq  \frac{2}{\mu} (B_F(p:r)+B_F(q:r))
\end{equation}

\end{description}

For mixtures with prescribed but different non-zero weighting, we can bound the likelihood ratio using $w^+=\max_i w_i\geq \frac{1}{k}$ and $w^-=\min_i w_i$.
When mixture weights are unknown, we can further discretize weights by increments of size $\delta$ ($O(1/\delta^k)$ such weight combinations, where each combination gives rise to a fixed weighting) and choose the initialization that yields the best likelihood. 

%
%

%
%
%

\section{Concluding remarks and discussion}\label{sec:concl}
Banerjee et al.~\cite{bregmankmeans-2005} proved that EM for learning exponential family mixtures amount to perform a dual Bregman soft clustering.
Based on the duality between exponential families and Bregman divergences, we proposed $k$-MLE, a Bregman hard clustering in disguise.
While $k$-MLE decreases monotonically the complete likelihood until it converges to a local minimum after a finite number of steps, EM monotonically decreases the expected complete likelihood and requires necessarily a prescribed stopping criterion.
Because $k$-MLE uses hard membership of observations, it fits the doubly stochastic process of sampling mixtures (for which soft EM  brings  mathematical convenience).  

Both $k$-MLE and EM are local search algorithm that requires to properly initialize the mixture parameters.
We described $k$-MLE++, a simple initialization procedure that builds on Bregman $k$-means++~\cite{BregmanClustering-2010} to probabilistically guarantee an initialization not too far from the global optimum (in case of known weights). 
While we use Lloyd $k$-means~\cite{LLoyd:1982} heuristic for minimizing the $k$-means loss, we can also choose other $k$-means heuristic to design a corresponding $k$-MLE. One possible choice is Hartigan's greedy swap~\cite{HartiganKmeans:2010} that can further improve the loss function when Lloyd's $k$-means is trapped into a local minimum. A local search technique such as Kanungo et al. swap~\cite{KanungoKmeans-2004} also guarantees a global $(9+\epsilon)$-approximation.

The MLE may yield degenerate situations when, say, one observation point is assigned to one component with weight close to one.
For example, the MLE of one point for the normal distribution is degenerate as $\sigma\rightarrow 0$ (and $w\rightarrow 1$)), and the likelihood function tends to infinity.
That is the unboundedness drawback of the MLE.
See~\cite{GMM-MLEPenalized:2001,GMMDegeneracy:2007} for further discussions on this topic including a penalization of the MLE to ensure boundedness.

Statistical mixtures with $k$ components are generative models of overall complexity $k-1+kD$, where $D$ is the order of the exponential family.
An interesting future direction would be to compare mixture models versus a {\em single} multi-modal exponential family~\cite{Cobb:1983:EMR} (with implicit log-normalizer $F$).
We did not address the model selection problem that consists in determining the appropriate number of components, nor the type of distribution family.
Although there exists many criteria like the Akaike Information Criterion (AIC), model selection is a difficult problem since some distributions exhibit the 
indivisibility property that makes the selection process unstable.
For example, a normal distribution can be interpreted as a sum of normal distributions: $\forall k\in\mathbb{N},\ N(\mu,\sigma^2) =  \sum_{i=1}^k N\left(\frac{\mu}{k},\frac{\sigma^2}{k}\right)$. 
From the practical point of view, it is better to overestimate $k$, and then perform mixture simplification using entropic clustering~\cite{jMEF-2010}.
Belkin and Sinha~\cite{PolynomialLearningDistr:2010} studied the polynomial complexity of learning a Gaussian mixture model.

We conclude by mentioning that it is still an active research topic to find good GMM learning algorithms in practice (e.g., see~the recent entropy-based algorithm~\cite{VBGMM:2012}).

%

\section*{Acknowledgments}

FN (5793b870) would like to thank Joris Geessels for an early  prototype in Python, Professor Richard Nock for stimulating discussions, 
and Professor Mario Tokoro and Professor Hiroaki Kitano for encouragements.

\appendix

\section{$k$-Means with per-cluster additively weighted Bregman divergence}\label{sec:kmeans}

$k$-Means clustering asks to minimize the   cost function $\kmeans(\X:\C)$ by partitioning input set $\X=\{x_1, ..., x_n\}$ into $k$ clusters using centers $\C=\{c_1, ..., c_k\}$, where

\begin{equation}\label{eq:kmeansEucl}
\kmeans(\X:\C) = \frac{1}{n} \sum_{i=1}^n \min_{j=1}^k \|x_i-c_j\|^2.
\end{equation}

There are several popular heuristics to minimize Eq.~\ref{eq:kmeansEucl} like Lloyd's batched method~\cite{Lloyd-1957} or Hartigan and Wong's swap technique~\cite{KmeansHartiganWong-1979}. Those iterative heuristics guarantee to decrease monotonically the $k$-means loss but can be trapped into a local minimum. In fact, solving for the global minimum $\kmeans^*(\X:\C)$ is NP-hard for general $k$ (even on the plane) and for $k=2$ and arbitrary dimension of datasets. Kanungo et al.~\cite{KanungoKmeans-2004} swap optimization technique guarantees a $(9+\epsilon)$-approximation factor, for any $\epsilon>0$.

Let us consider  an additively weighted Bregman divergence $B_{F_i,m_i}$ per cluster as follows:

\begin{equation}
B_{F_i,m_i}(p:q) = B_{F_i}(p:q) + m_i, 
\end{equation}
with $m_i$ denoting the additive mass attached to a cluster center\footnote{In this paper, we have $m_i\geq 0$ by choosing $m_i=-\log w_i$ for $w_i<1$, but this is not required.}, and $B_{F_i}$ the Bregman divergence  induced by   the Bregman generator $F_i$ defined  by
\begin{equation}
B_{F_i}(p:q) = F_i(p)-F_i(q)-\inner{p-q}{\nabla F_i(q)},
\end{equation}

\begin{remark}
For $k$-MLE, we consider all component distributions of the same exponential family $E_F$, and therefore all $F_i=F^*$'s are identical. We could have also considered different exponential families for the components but this would have burdened the paper with additional notations although it is of practical interest.
For example, for the case of the multivariate Gaussian family, we can split the vector parameter part from the matrix parameter part, and write 
$F({\theta_v}_i,{\theta_M}_i)=F_{{\theta_M}_i}({\theta_v}_i) = F_i({\theta_v}_i)$. 
\end{remark}

Let us extend the Bregman batched Lloyd's $k$-means clustering~\cite{bregmankmeans-2005} by considering the generalized $k$-means clustering loss function for a data set 
$\Y=\{y_1, ..., y_n\}$ and a set $\C$ of $k$ cluster centers $\C=\{c_1, ..., c_k\}$:

\begin{equation}
\kmeans(\Y,\C) = \min_{c_1, ..., c_k} \frac{1}{n} \sum_{i=1}^n \min_{j=1}^k D_i(y_i : c_j).
\end{equation}

Let us prove that the center-based Lloyd's $k$-means clustering algorithm monotonically decreases this loss function, and terminates after a finite number of iterations into a local optimum.

\begin{itemize}
\item 

When $k=1$, the minimizer of $\kmeans(\Y,\C=\{c_1\})$ is the center of mass (always independent of the Bregman generator):
\begin{equation}
c_1=\frac{1}{n}\sum_{i=1}^n y_i =\bar y,
\end{equation}
and the Bregman information~\cite{bregmankmeans-2005} is defined as the minimal $1$-means loss function:

\begin{eqnarray}
\kmeans_{F_1,m_1}(\Y,\{c_1\}) &=& I_{F_1}(\Y)\\
&=& \frac{1}{n} \sum_{i=1}^n F_1(y_i)- F_1(\bar y) + m_1,\\
&=& m_1+J_{F_1}(\Y),
\end{eqnarray}
where $\bar y=\frac{1}{n}\sum_{i=1}^n y_i$ and
\begin{equation}
J_{F_1}(y_1, ..., y_n)=\frac{1}{n} \sum_{i=1}^n F_1(y_i)- F_1(\bar y) \geq 0,
\end{equation}
denotes the Jensen diversity index~\cite{2009-BregmanCentroids-TIT}.

\item When $k\geq 2$, let $c_i^{(t)}$  denote the cluster center of the $i$-th cluster $\C_i^{(t)}\subset\Y$  of the partition $\Y=\cup_{i=1}^k \C_i^{(t)}$ at the $t^{\mathrm{th}}$ iteration. 
The generalized additively weighted Bregman $k$-means loss function can be rewritten as

\begin{equation}
\kmeans_{F,m}(\C_1^{(t)}, ...,\C_k^{(t)} : c_1^{(t)}, ..., c_k^{(t)}) = \frac{1}{n} \sum_{i=1}^k \sum_{y\in\C_i^{(t)}} B_{F_i,m_i}(y : c_i).
\end{equation}

Since the assignment step allocates  $y_i$ to their closest cluster center $\argmin_{j=1}^k B_{F_i,m_i}(y_i : c_j)$,   we have

\begin{equation}
\kmeans_{F,m}(\C_1^{(t+1)}, ...,\C_k^{(t+1)} : c_1^{(t)}, ..., c_k^{(t)})  \leq \kmeans_{F,m}(\C_1^{(t)}, ...,\C_k^{(t)} : c_1^{(t)}, ..., c_k^{(t)}).
\end{equation}

Since the center relocation minimizes the average additively weighted divergence, we have

\begin{equation}
\kmeans_{F,m}(\C_1^{(t+1)}, ...,\C_k^{(t+1)} : c_1^{(t+1)}, ..., c_k^{(t+1)})  \leq \kmeans_{F,m}(\C_1^{(t+1)}, ...,\C_k^{(t+1)}; c_1^{(t)}, ..., c_k^{(t)}).
\end{equation}

By iterating the assignment-relocation steps of $k$-means, and cascading the inequalities by transitivity, we get

\begin{equation}
\kmeans_{F,m}(\C_1^{(t+1)}, ...,\C_k^{(t+1)} : c_1^{(t+1)}, ..., c_k^{(t+1)}) \leq \kmeans_{F,m}(\C_1^{(t)}, ...,\C_k^{(t)} : c_1^{(t)}, ..., c_k^{(t)})
\end{equation}

Since the loss function is trivially lower bounded by $\frac{1}{n} \min_{i=1}^k m_i$ (and therefore always positive when all $m_i\geq 0$), we conclude that the generalized Bregman $k$-means converge to a local optimum, after a finite number\footnote{We cannot repeat twice a partition.} of iterations.

Furthermore, the loss function can be expressed as

\begin{eqnarray}
\kmeans_{F,m}(\C_1, ...,\C_k : c_1, ..., c_k) &=& \frac{1}{n} \sum_{i=1}^k \sum_{y\in\C_i} B_{F_i,m_i}(y : c_i),\\
& = & \sum_{i=1}^k  w_i J_{F_i }(\C_i) + \sum_{i=1}^k w_i m_i,\label{eq:kmeans}
\end{eqnarray}
with  $J_{F_i}(\C_i)=\frac{1}{|\C_i|} \sum_{y\in\C_i}^n F_i(y)- F_i(c_i)\geq 0$ (and $c_i=\frac{\sum_{y\in\C_i} y}{|\C_i|}$), and $w_i=\frac{|\C_i|}{n}$ for all $i\in\{1, ..., k\}$, the cluster relative weights.
\end{itemize}

When all $F_i$ are identical to some generator $F$, we have the following loss function:

\begin{equation}
\kmeans_{F,m} = \sum_{i=1}^k  w_i J_{F }(\C_i) + \sum_{i=1}^k w_i m_i
\end{equation}

The celebrated $k$-means of Lloyd~\cite{Lloyd-1957} minimizes the weighted within-cluster variances (for the Bregman quadratic generator $F(x)=\inner{x}{x}$ inducing the squared Euclidean distance error) as shown in Eq.~\ref{eq:kmeans}, 
with Bregman information:
\begin{eqnarray}
J_F(\Y) &=& \sum_{y\in\Y} \frac{1}{|\Y|} \| y-\bar y\|^2,\\
&=& \sum_{y\in\Y} \frac{1}{|\Y|} \inner{y-\bar y}{y-\bar y},\\
&=& \sum_{y\in\Y} \frac{1}{|\Y|} (\inner{y}{y}-2\inner{\bar y}{y}-\inner{\bar y}{\bar y}),\\
&=& \sum_{y\in\Y} \frac{1}{|\Y|} \inner{y}{y} -2\Inner{\bar y}{\underbrace{\sum_{y\in\Y} \frac{1}{|\Y|} y}_{\bar y}}-\inner{\bar y}{\bar y},\\
& =& \frac{1}{|\Y|}\sum_{y\in\Y}\inner{y}{y}  - \inner{\bar y}{\bar y} = J_F(\Y),
\end{eqnarray}
the variance.
When all cluster generators are identical and have no mass, it is shown by Banerjee et al.~\cite{bregmankmeans-2005} that the loss function can be equivalently rewritten as:

\begin{eqnarray}
\kmeans_F(\P : \C)  &=& J_F(\P)-J_F(\C) = \sum_{i=1}^k w_i J_F(\C_i),\\
&=& I_F(\P)-I_F(\C)
\end{eqnarray} 

\begin{remark} 
Note that we always have $\bar c=\bar y$.
That is, the centroid $\bar y$ of set $\Y$ is equal to the barycenter $\bar c$ of the cluster centers $\C$ (with weights taken as the relative proportion of points falling within the clusters. 
\end{remark}

\begin{remark}
A multiplicatively weighted Bregman divergence $m_iB_{F_i}$ is mathematically equivalent to a Bregman divergence $B_{m_iF_i}$ for generator $m_iF_i$, provided that $m_i>0$.
\end{remark}

As underlined in this proof, Lloyd's $k$-means~\cite{Lloyd-1957} assignment-center relocation loop is a generic algorithm that extends  to arbitrary divergences $D_i$ guaranteeing unique average divergence minimizers, and the assignment/relocation process ensures that the associated $k$-means loss function decreases monotonically.
Teboulle studied~\cite{TeboulleCluster:2007} generic center-based clustering optimization methods.  
It is however difficult to reach the global minimum since $k$-means is NP-hard, even when data set $\Y$ lies on the plane~\cite{kmeans-expiter2D:2011} for arbitrary $k$. 
In the worst case, $k$-means may take an exponential number of iterations to converge~\cite{kmeans-expiter2D:2011}, even on the plane.

\section{Dual parameterization of the multivariate Gaussian (MVN) family}\label{sec:mvn}
Let us explicit the dual $\theta$-natural and $\eta$-moment parameterizations of the family of multivariate Gaussians.
Consider the multivariate Gaussian probability density parameterized by a mean vector $\lambda_v=\mu$ and a covariance matrix $\lambda_M=\Sigma$.

\begin{eqnarray} 
p(x;\lambda)&=&\frac{1}{(2\pi)^{\frac{d}{2}}\sqrt{|\lambda_M|}}e^{- \frac{1}{2}(x-\lambda_v)^T \lambda_M^{-1} (x-\lambda_v)},\\
&=& \exp \left( -\frac{1}{2} x^T\lambda_M^{-1} x +\lambda_v^T\lambda_M^{-1}x-\frac{1}{2}\lambda_v^T\lambda_M^{-1}\lambda_v-\frac{d}{2}\log 2\pi-\frac{1}{2}\log|\lambda_M| \right),
\end{eqnarray}
where the usual parameter is $\lambda=(\lambda_v,\lambda_M)=(\mu,\Sigma)$.
Using the matrix cyclic trace property $-\frac{1}{2} x^T\lambda_M^{-1} x=\tr(-\frac{1}{2}x x^T \lambda_M^{-1})$ and the fact that $(\lambda_M^{-1})^T=\lambda_M^{-1}$, we rewrite the density as follows:

\begin{equation}
p(x;\lambda)=\exp \left( \inner{x}{\lambda_M^{-1}\lambda_v}+\inner{-\frac{1}{2}x x^T}{\lambda_M^{-1}}  - \left (\frac{1}{2}\lambda_v^T\lambda_M^{-1}\lambda_v+\frac{d}{2}\log 2\pi+\frac{1}{2}\log|\lambda_M| \right)  \right),
\end{equation}
where the inner product of vector is $\inner{v_1}{v_2}=v_1^T v_2$ and the inner product of matrices is $\inner{M_1}{M_2}=\tr(M_1^T M_2)$.
Thus we define the following canonical terms:

\begin{itemize}
\item sufficient statistics: $t(x)=(x,-\frac{1}{2} x x^T)$, 
\item auxiliary carrier measure: $k(x)=0$,
\item natural parameter: $\theta=(\theta_v,\theta_M)=(\lambda_M^{-1}\lambda_v,\lambda_M^{-1})$.

\item log-normalizer expressed in the $\lambda$-coordinate system:

\begin{equation}
F(\lambda)=\frac{1}{2}\lambda_v^T\lambda_M^{-1}\lambda_v+\frac{d}{2}\log 2\pi+\frac{1}{2}\log|\lambda_M|
\end{equation}

Since $\lambda_v=\theta_M^{-1}\theta_v$ (and $\lambda_v^T=\theta_v^T\theta_M^{-1}$) and $\log|\lambda_M|=-\log|\theta_M|$, we express the log-normalizer in the $\theta$-coordinate system as follows:

\begin{equation}
F(\theta)= \frac{1}{2}\theta_v^T\theta_M^{-1}\theta_v -\frac{1}{2}\log|\theta_M|+\frac{d}{2}\log 2\pi
\end{equation}

\end{itemize} 

Since the derivative of the log determinant of a symmetric matrix is $\nabla_X \log|X|=X^{-1}$ and the derivative of an inverse matrix trace~\cite{MatrixCookbook}:

\begin{equation}
\nabla_X \tr(A X^{-1} B)=-(X^{-1} BA X^{-1})^{T}
\end{equation}
(applied to $ \frac{1}{2}\tr(\theta_v^T\theta_M^{-1}\theta_v) = -\frac{1}{2}(\theta_M^{-1}\theta_v\theta_v^T\theta_M^{-1})$), we
calculate the gradient $\nabla F$ of the log-normalizer as

\begin{equation}
\nabla F(\theta) = (\nabla_{\theta_v}F(\theta),\nabla_{\theta_M} F(\theta))
\end{equation}
with

\begin{eqnarray}
\eta_v&=&\nabla_{\theta_v}F(\theta)=\theta_M^{-1}\theta_v,\\
&=& E[x]=\mu,\\
\eta_M&=&\nabla_{\theta_M} F(\theta)= -\frac{1}{2} (\theta_M^{-1}\theta_v)(\theta_M^{-1}\theta_v)^T -\frac{1}{2}\theta_M^{-1},\\
&=& E\left[-\frac{1}{2}xx^T\right]=-\frac{1}{2}(\mu\mu^T+\Sigma),
\end{eqnarray}
where $\eta=\nabla F(\theta)=(\eta_v,\eta_M)$ denotes the dual moment parameterization of the Gaussian.

It follows that the Kullback-Leibler divergence of two multivariate Gaussians is

\begin{eqnarray}
\KL(p(x;\lambda_1):p(x;\lambda_2)) &=& B_F(\theta_2:\theta_1),\\
&=& \frac{1}{2}\left(\tr(\Sigma_2^{-1}\Sigma_1)-\log |\Sigma_1\Sigma_2^{-1}|+(\mu_2-\mu_1)^T \Sigma_2^{-1} (\mu_2-\mu_1)\right).
\end{eqnarray}

Note that the Kullback-Leibler divergence of  multivariate Gaussian distributions~\cite{GaussianClustering:2006} can be decomposed as the sum of a Burg matrix divergence (Eq.~\ref{eq:burg}) with a squared Mahalanobis distance (Eq.~\ref{eq:Mah}) (both being Bregman divergences):

\begin{eqnarray}
\KL(p_F(x|\mu_1,\Sigma_1) : p_F(x|\mu_2,\Sigma_2) &=& \frac{1}{2}\left(\tr(\Sigma_2^{-1}\Sigma_1)-\log |\Sigma_1\Sigma_2^{-1}|+(\mu_2-\mu_1)^T \Sigma_2^{-1} (\mu_2-\mu_1)\right)\\
&= & \frac{1}{2}B(\Sigma_1,\Sigma_2)+ \frac{1}{2} M_{\Sigma_2^{-1}}(\mu_1,\mu_2),
\end{eqnarray}
with
\begin{eqnarray}
B(\Sigma_1 : \Sigma_2) &=& \tr(\Sigma_1\Sigma_2^{-1})-\log|\Sigma_1 \Sigma_2^{-1}|-d,\\ \label{eq:burg}
M_{\Sigma_2^{-1}}(\mu_1,\mu_2) &= & (\mu_1-\mu_2)^T \Sigma_2^{-1} (\mu_1-\mu_2).  \label{eq:Mah}
\end{eqnarray}

To compute the functional inverse of the gradient, we write:

\begin{equation}
\theta=\nabla F^{-1}(\eta)=\nabla F^*(\eta).
\end{equation}

Since $\eta_M=-\frac{1}{2}(\eta_v\eta_v^T+\theta_M^{-1})$, we have:

\begin{eqnarray}
\theta_M &=& (-2\eta_M-\eta_v\eta_v^T)^{-1},\\
\theta_v &=& (-2\eta_M-\eta_v\eta_v^T)^{-1}\eta_v.
\end{eqnarray}

Finally, we get the Legendre convex conjugate $F^*(\eta)$ as:

\begin{eqnarray}
F^*(\eta)&=&\inner{\nabla F^*(\eta)}{\eta}-F(\nabla F^*(\eta)),\\
&=&  -\frac{1}{2}\log (1+2\eta_v^T \eta_M^{-1} \eta_v) -\frac{1}{2}\log|-\eta_M|-\frac{d}{2}\log (\pi e).
\end{eqnarray}

\section{$k$-MLE for Gaussian Mixture Models (GMMs)}

We explicit $k$-MLE for Gaussian mixture models on the usual $(\mu,\Sigma)$ parameters in Algorithm~\ref{algo:kmlegmm}.

\begin{algo}
\caption{$k$-MLE for learning a GMM.\label{algo:kmlegmm}}

\underline{Input}:\\
\begin{tabular}{lll}
$X$ &:&  a set of $n$ independent and identically distributed distinct observations: $X=\{x_1, ..., x_n\}$\\
$k$ &:& number of clusters\\
\end{tabular}

\begin{itemize}
\item 0. {\bf Initialization}: 

\begin{itemize}
\item Calculate global mean $\bar\mu$ and global covariance matrix $\bar\Sigma$:
\begin{eqnarray*}
\bar\mu &=& \frac{1}{n} \sum_{i=1}^k x_i, \\
\bar\Sigma&=&\frac{1}{n}  \sum_{i=1}^k x_ix_i^T - \bar\mu\bar\mu^T
\end{eqnarray*}

\item $\forall i\in\{1, ...,k\}$, initialize the $i$th seed as $(\mu_i=x_i, \Sigma_i=\bar\Sigma)$.
\end{itemize}

\item 1. {\bf Assignment}: 

$$
\forall i\in\{1, ...,n\}, z_i=\argmin_{j=1}^k M_{\Sigma_i^{-1}}(x-\mu_i,x-\mu_i)+\log |\Sigma_i|-2\log w_i
$$
with $M_{\Sigma_i^{-1}}(x-\mu_i,x-\mu_i)$ the squared Mahalanobis distance: $M_Q(x,y)=(x-y)^T Q(x-y)$.

Let $\C_i=\{x_j | z_j=i\}, \forall i\in\{1, ..., k\}$ be the cluster partition: $X=\cup_{i=1}^k \C_i$.\\
(Anisotropic Voronoi diagram~\cite{AnisotropicVoronoiDiagram:2003})

\item 2. {\bf Update the parameters}: 

$$
\forall i\in\{1, ..., k\}, \mu_i=\frac{1}{|\C_i|}\sum_{x\in\C_i} x, \Sigma_i=\frac{1}{|\C_i|}\sum_{x\in\C_i} xx^T -\mu_i\mu_i^T
$$

{\bf Goto step~1} unless local convergence of the complete likelihood is reached.

\item 3. {\bf Update the mixture weights}: 
$\forall i\in\{1, ..., k\}, w_i=\frac{1}{n}|\C_i|$.\\
{\bf Goto step~1} unless local convergence of the complete likelihood is reached.
\end{itemize}

\end{algo}

The $k$-MLE++ initialization for the GMM is reported in Algorithm~\ref{algo:kmleppgmm}.

\begin{algo}
\caption{$k$-MLE for GMM: \label{algo:kmleppgmm}}
 
\begin{itemize}

\item Choose first seed $\C=\{y_l\}$, for $l$ uniformly random in $\{1, ..., n\}$.

\item For $i\leftarrow 2$ to $k$

\begin{itemize}
\item Choose $c_i=(\mu_i,\Sigma_i)$ with probability 

$$
\frac{B_{F*}(c_i:\C)}{\sum_{i=1}^n B_{F*}(y_i:\C)}  = \frac{B_{F^*}(\Y:\C)}{\kmeans_{F^*}(\Y:\C)},
$$
where $B_{F^*}(c:\P)=\min_{p\in\P} B_{F^*}(c:p)$.

$$
F^*(\mu,\Sigma) = -\frac{1}{2}\log \left(1-\mu^T(\mu\mu^T+\Sigma)^{-1}\mu\right)-\frac{1}{2}\log|\mu^T\mu+\Sigma|-\frac{d}{2}\log 2\pi-d
$$
 
\item Add selected seed to the initialization seed set: $\C\leftarrow \C\cup \{c_i\}$.

\end{itemize}
\end{itemize}

\end{algo}

\section{Rayleigh Mixture Models (RMMs)}

We instantiate the soft Bregman EM, hard EM, $k$-MLE, and $k$-MLE++ for the Rayleigh distributions, a sub-family of Weibull distributions.

A Rayleigh distribution has probability density $\frac{x}{\sigma^2}e^{-\frac{x^2}{2\sigma^2}}$ where $\sigma\in\mathbb{R}^+$ denotes the {\it mode} of the distribution, and $x\in\mathbf{X}=\mathbb{R}^+$ the support. 
The Rayleigh distributions form a $1$-order univariate exponential family ($D=d=1$).
Re-writing the density in the canonical form $e^{ -\frac{x^2}{2\sigma^2}+\log x-2\log\sigma}$, we deduce that
$t(x)=x^2$, $\theta=-\frac{1}{2\sigma^2}$, $k(x)=\log x$, and $F(\sigma^2)=\log\sigma^2=\log -\frac{1}{2\theta}=-\log (-2\theta)=F(\theta)$.
Thus $\nabla F(\theta)=-\frac{1}{\theta}=\eta$ and $F^*(\eta)=\inner{\theta}{\eta}-F(\theta)=-1+\log\frac{2}{\eta}$.
The natural parameter space is $\thetaspace=\mathbb{R}^-$ and the moment parameter space is $\etaspace=\mathbb{R}^+$ (with $\eta=2\sigma^2$).
We check that conjugate gradients are reciprocal of each other since $\nabla F^*(\eta)=-\frac{1}{\eta}=\theta$, and we have 
$\nabla^2 F(\theta) \nabla^2 G(\eta) =\frac{1}{\theta^2} \frac{1}{\eta^2}  = 1$ (i.e, dually orthogonal coordinate system) with $\nabla^2 F(\theta)=\frac{1}{\theta^2}$ and $\nabla^2 F^*(\eta)=\frac{1}{\eta^2}$.

Rayleigh mixtures are often used in ultrasound imageries~\cite{RMM:2011}.

\subsection{EM as a Soft Bregman clustering algorithm}

Following Banerjee et al.~\cite{bregmankmeans-2005}, we instantiate the Bregman soft clustering for the convex conjugate $F^*(\eta)=-1+\log \frac{2}{\eta}$, $t(x)=x^2$ and $\eta=2\sigma^2$. The Rayleigh density expressed in the $\eta$-parameterization yields
$p(x;\sigma)=p(x;\eta)=\frac{2x}{\eta}e^{-\frac{2x^2}{\eta}}$.

\begin{description}

\item[Expectation.] Soft membership for all observations $x_1, ..., x_n$:

\begin{eqnarray}
\forall 1\leq i\leq n, 1\leq j\leq k,\ 
w_{i,j}  &=& \frac{w_j p(x_i;\theta_j)}{\sum_{l=1}^k w_{l} p(x_i;\theta_{l})},
\end{eqnarray}
(We can use any of the equivalent $\sigma$, $\theta$ or $\eta$ parameterizations for calculating the densities.)

\item[Maximization.] Barycenter in the moment parameterization:

\begin{eqnarray}
\forall  1\leq j\leq k,\
\eta_j &=& \frac{\sum_{i=1}^n w_{i,j} t(x_i)}{\sum_{l=1}^n w_{l,j} },\\
\sigma_j &=& \sqrt{\frac{1}{2} \frac{\sum_{i=1}^n w_{i,j} x_i^2}{\sum_{l=1}^n w_{l,j} }}
\end{eqnarray}

\end{description}

\subsection{$k$-Maximum Likelihood Estimators}

The associated Bregman divergence for the convex conjugate generator of the Rayleigh distribution log-normalizer is 

\begin{eqnarray}
B_{F^*}(\eta_1 : \eta_2) &=& F^*(\eta_1) - F^*(\eta_2) - \inner{\eta_1-\eta_2}{\nabla F^*(\eta_2)},\\
&=& -1+\log \frac{2}{\eta_1} + 1-\log \frac{2}{\eta_2} - (\eta_1-\eta_2)(-1/\eta_2),\\
&=& \frac{\eta_1}{\eta_2} + \log \frac{\eta_2}{\eta_1} - 1,\\
&=& \IS(\eta_1:\eta_2)
\end{eqnarray}

This is the Itakura-Saito divergence IS (indeed, $F^*$ is equivalent modulo affine terms to $-\log \eta$, the Burg entropy).

\begin{description}

\item[1. Hard assignment.] 

$$
\forall 1\leq i\leq n, 
z_i= \argmin_{1\leq j\leq k}  \IS(x_i^2 : \eta_j) - \log w_j
$$

Voronoi partition into clusters:

$$
\forall 1\leq j\leq k, 
\C_j = \{  x_i\ |\  \IS( x_i^2 : \eta_j) - \log w_j  \leq \IS( x_i^2 : \eta_l) - \log w_l\forall l\not=j \}
$$

\item[2. $\eta$-parameter update.]

$$
\forall 1\leq j\leq k, \eta_j \leftarrow \frac{1}{|\C_j|} \sum_{x\in\C_j} x^2
$$

$$
\forall 1\leq j\leq k, \sigma_j=\sqrt{\frac{1}{2}\eta_j}
$$

Go to 1. until (local) convergence is met.

\item[weight update.]

$$
\forall 1\leq j\leq k, w_j=\frac{|\C_j|}{n}
$$

Go to 1. until (local) convergence is met.

\end{description}

Note that $k$-MLE does also model selection as it may decrease the number of clusters in order to improve the complete log-likelihood.
If initialization is performed using random point and uniform weighting, the first iteration ensures that all Voronoi cells are non-empty.

\subsection{$k$-MLE++}

A good initialization for Rayleigh mixture models is done as follows:
Compute the order statistics for the $\frac{n}{k}, \frac{2n}{k}, \frac{(k-1)n}{k}$-th elements (in overall $O(n\log k)$-time).
Those pivot elements split the set $\X$ into $k$ groups $\X_1, ..., \X_k$ of size $\frac{n}{k}$, on which we estimate the MLEs.

The $k$-MLE++ initialization is built from the Itakura-Saito divergence:
$$
\IS(\eta_1:\eta_2) = \frac{\eta_1}{\eta_2} + \log \frac{\eta_2}{\eta_1} - 1
$$ 

k-MLE++:
\begin{itemize}

\item Choose first seed $\C=\{y_l\}$, for $l$ uniformly random in $\{1, ..., n\}$.

\item For $i\leftarrow 2$ to $k$

\begin{itemize}

\item Choose $c_i\in y_1=x_1^2, ..., y_n=x_n^2$ with probability 

$$
\frac{\IS(c_i:\C)}{\sum_{i=1}^n \IS(y_i:\C)}
$$

\item Add selected seed to the initialization seed set: $\C\leftarrow \C\cup \{c_i\}$.

\end{itemize}

\end{itemize}

%
%
%
%
%
%
%
%
%
%
%
%
%
%
%

\newpage
\section{Notations}\label{sec:notations}

\begin{supertabular}{ll}
\underline{Exponential family}:\\
$\inner{x}{y}$ & inner product (e.g., $x^\top y$ for vectors, $\tr (Y^\top X)$ for matrices)\\
$p_F(x;\theta) = e^{\inner{t(x)}{\theta}-F(\theta)+k(x)}$ & Exponential distribution parameterized using the $\theta$-coordinate system\\
$\mathbb{X}$ & support of the distribution family ($\{x\ |\ p_F(x;\theta)>0 \}$)\\ 
$d$ & dimension of the support $\mathbb{X}$ (univariate versus multivariate)\\
$D$ & dimension of the natural parameter space\\
& (uniparameter versus multiparameter)\\
$t(x)$ & sufficient statistic ($\hat\eta=\frac{1}{n}\sum_{i=1}^n t(x_i)$)\\
$k(x)$ & auxiliary carrier term\\
$F$ & log-normalizer, log-Laplace, cumulant function ($F: \thetaspace\rightarrow \mathbb{R}$) \\
$\nabla F$ & gradient of the log-normalizer (for moment $\eta$-parameterization)\\
$\nabla^2 F$ & Hessian of the log-normalizer\\
& (Fisher information matrix, SPD: $\nabla^2 F(\theta)\succ 0$)\\
$F^*$ & Legendre convex conjugate\\
\underline{Distribution parameterization}:\\
$\theta$ & canonical natural parameter\\
$\thetaspace$ & natural parameter space\\
$\eta$ & canonical moment parameter\\
$\etaspace$ & moment parameter space\\
$\lambda$ & usual parameter\\
$\lambdaspace$ & usual parameter space\\
$p_F(x;\lambda)$ & density or mass function using the usual $\lambda$-parameterization\\
$p_F(x;\eta)$ & density or mass function using the usual moment parameterization\\
\underline{Mixture}:\\
$m$ & mixture model\\
$\Delta_k$ & closed probability $(d-1)$-dimensional simplex\\
$H(w)$ & Shannon entropy $-\sum_{i=1}^d w_i\log w_i$ (with $0\log 0=0$ by convention)\\
$H^\times(p:q)$ & Shannon cross-entropy $-\sum_{i=1}^d p\log q$\\
$w_i$ & mixture weights (positive such that $\sum_{i=1}^k w_i=1$)\\
$\theta_i$ & mixture component natural parameters\\
$\eta_i$ & mixture component moment parameters\\
$\tilde m$ & estimated mixture\\
$k$ & number of mixture components\\
$\Omega$ & mixture parameters\\
\underline{Clustering}:\\
$\X=\{x_1, ..., x_n\}$ & sample (observation) set\\
$|\X|, |\C|$ & cardinality of sets: $n$ for the observations, $k$ for the cluster centers\\
$z_1, ..., z_n$ & Hidden component labels\\
$\Y=\{y_1=t(x_1), ..., y_n=t(x_n)\}$ & sample sufficient statistic set\\
$L(x_1, ..., x_n;\theta)$ & likelihood function\\
$\hat\theta,\hat\eta,\hat\lambda$ & maximum likelihood estimates\\
$w_{i,j}$ & soft weight for $x_i$ in cluster/component $\C_j$ ($w_j, \theta_j$)\\
$i$ & index on the sample set $x_1, ..., x_i, ..., x_n$\\
$j$ & index on the mixture parameter set $\theta_1, ...,\theta_j, ...,\theta_k$\\
$\C$ & cluster partition\\
$c_1, ..., c_k$ & cluster centers\\
$\alpha_1, ..., \alpha_k$ & cluster proportion size\\
$B_F$ & Bregman divergence with generator $F$:\\
 & 
 \begin{minipage}{10cm}
 \begin{eqnarray*}
 B_F(\theta_2,\theta_1) &=& \KL(p_F(x:\theta_1) : p_F(x:\theta_2))\\
 &=& B_{F^*}(\eta_1,\eta_2)\\
 &=& F(\theta_2)+F^*(\eta_1)-\inner{\eta_1}{\theta_2}
 \end{eqnarray*}
 \end{minipage}\\
$J_F$ & Jensen diversity index:\\
& $J_F(p_1, ..., p_n; w_1, ..., w_n)= \sum_{i=1}^n w_i F(p_i) - F(\sum_{i=1}^n w_ip_i) \geq 0$\\
\underline{Evaluation criteria}:\\
$\bar l_F$ & average incomplete log-likelihood:\\
& $\bar l_F(x_1, ..., x_n)=\frac{1}{n} \sum_{i=1}^n \log \sum_{j=1}^k w_j p_F(x_i;\theta_j)$\\
$\bar l'_F$ & average complete log-likelihood\\
& $\bar l'_F(x_1, ..., x_n)=\frac{1}{n} \sum_{i=1}^n \log  w_{z_i} p_F(x_i;\theta_{z_i})$\\
$\bar L_F$ & geometric average incomplete likelihood:\\
& $\bar L_F(x_1, ..., x_n)=e^{\bar l_F(x_1, ..., x_n)}$ \\
$\bar L'_F$ & geometric average complete likelihood:\\
 & $\bar L'_F(x_1, ..., x_n)=e^{\bar l'_F(x_1, ..., x_n)}$ \\
$\kmeans_F$ & average $k$-means loss function (average divergence to the closest center)\\
& \begin{minipage}{10cm}
\begin{eqnarray*}
\kmeans_F(\X,\C) &=& \frac{1}{n} \sum_{i=1}^n B_F(x_i : \C)\\
&=& \frac{1}{n} \sum_{j=1}^k \sum_{x\in\C_j} B_F(x : c_j)\\
&=& \sum_{j=1}^k w_j J_F(\C_j)\\
&=& J_F(\X)-J_F(\C)
\end{eqnarray*}
\end{minipage}\\
$\kmeans_{F,m}$ & average $k$-means loss function with respect to additive Bregman divergences\\
\end{supertabular}

\newpage


\end{document}